\documentclass{article}
\usepackage{ijcai24}

\usepackage{times}
\usepackage{soul}
\usepackage{url}
\usepackage[utf8]{inputenc}
\usepackage[small]{caption}
\usepackage{graphicx}
\usepackage{amsmath}
\usepackage{amsthm}
\usepackage{booktabs}
\usepackage{algorithm}
\usepackage{algorithmic}
\usepackage[switch]{lineno}

\usepackage{latexsym}
\usepackage{amssymb}
\usepackage{tabularx}
\usepackage{multirow}
\usepackage{makecell} 
\usepackage{color}
\usepackage[colorlinks,
linkcolor=blue,       
anchorcolor=blue,  
citecolor=blue,        
urlcolor=blue
]{hyperref}
\usepackage{supertabular}
\usepackage{float}
\usepackage{stfloats}
\usepackage{cite}
\usepackage{array}
\usepackage{bbding}
\usepackage[misc]{ifsym}

\title{Integrating Hierarchical Semantic into Iterative Generation Model \\ for Entailment Tree Explanation}

\author{
Qin Wang$^1$
\and
Jianzhou Feng$^1$\and
Yiming Xu$^1$\\
\affiliations
$^1$School of Information Science and Engineering, Yanshan University\\
\emails
veyc@stumail.ysu.edu.cn,
fjzwxh@ysu.edu.cn
}

\begin{document}

\maketitle

\begin{abstract}
    Manifestly and logically displaying the line of reasoning from evidence to answer is significant to explainable question answering (QA). The entailment tree exhibits the lines structurally, which is different from the \textit{self-explanation} principle in large-scale language models. Existing methods rarely consider the semantic association of sentences between and within hierarchies within the tree structure, which is prone to apparent mistakes in combinations. In this work, we propose an architecture of integrating the Hierarchical Semantics of sentences under the framework of Controller-Generator (HiSCG) to explain answers. The HiSCG designs a hierarchical mapping between hypotheses and facts, discriminates the facts involved in tree constructions, and optimizes single-step entailments. To the best of our knowledge, We are the first to notice hierarchical semantics of sentences between the same layer and adjacent layers to yield improvements. The proposed method achieves comparable performance on all three settings of the EntailmentBank dataset. The generalization results on two out-of-domain datasets also demonstrate the effectiveness of our method.
\end{abstract}

\section{Introduction}
Plausible explanations are fundamental to artificial intelligence, which underpin the results predicted by models and make an important contribution to human understanding. The reasoning capabilities of Large-scale Language Models (LLMs) elicited by Chain of Thought (CoT) prompts help boost the quality of the explanations in natural language~\shortcite{NEURIPS2022_9d560961,NEURIPS2022_8bb0d291, qinChatGPTGeneralPurposeNatural2023}. But they still suffer from hallucinations on many tasks~\cite{jiSurveyHallucinationNatural2023,leeMathematicalInvestigationHallucination2023}. Credible explanation generation has undoubtedly become a principal aspect of the implementation of artificial intelligence.

\begin{figure}[!t]
	\centering
	\includegraphics[width=3.2in,height=3.0in]{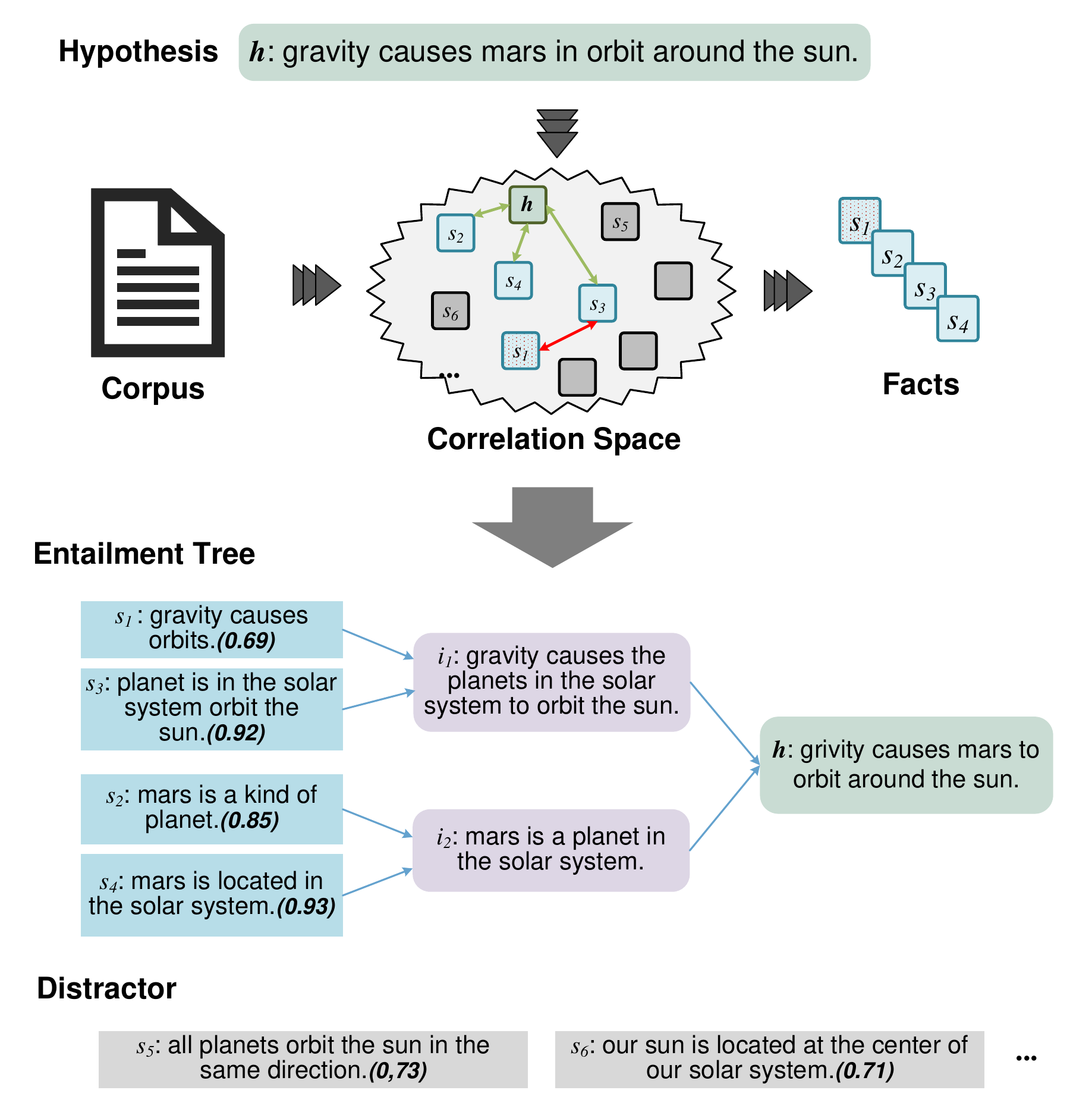}
	\caption {The task of explaining a hypothesis from premises with an entailment tree. A hypothesis is comprised of a question and an answer. Premises (blue) are obtained from the provided corpus (or given directly) with distractors (grey). Premises act as leaf nodes in the entailment tree which contains intermediate nodes (purple) and hypothesis as the root node (red). The values in parentheses are calculated by \textit{Sentence-BERT}~\cite{reimers-gurevych-2019-sentence} using cosine similarity between facts and hypotheses.} 
	\label{question_describe}
\end{figure}
We follow \cite{dalvi-etal-2021-explaining} to adopt the form of entailment trees to generate credible explanations, which employ verified facts and standardized interpretation steps, to present the decision-making process transparently and controllably. As illustrated in Fig.~\ref{question_describe}, for a given hypothesis, the tree will explain why the hypothesis is correct. Facts in the corpus will become the leaf nodes, and the hypothesis is obtained by multi-entailment steps. The popular approach employs sequence-to-sequence models to output linearized entailment trees in a time~\cite{neves-ribeiro-etal-2022-entailment, dalvi-etal-2021-explaining}. However, this approach is not conducive to error analysis. Another existing approach is based on the controller-generator framework, which iteratively generates single entailment steps through two modules that operate independently~\cite{hong-etal-2022-metgen, liu-etal-2022-rlet}. The controller selects facts in each step, while the generator generates the conclusion after combining those facts (also known as the premises). In comparison, the controller-generator framework-based approaches have a rigorous derivation principle. 

Nevertheless, most previous methods use semantic similarity to filter premises. As shown in Fig.~\ref{question_describe}, the distractors (i.e., $s_5$,$s_6$) have higher semantic similarity to the hypothesis compared to $s_1$, but the hypothesis cannot be derived from it. Intuitively, the hierarchy serves as the basis of trees, and there should be hierarchical semantic relationships between sentences in the tree. If we can also hierarchically organize the flat facts in the corpus, it will be beneficial to subsequent operations. Therefore, we try to adjust the sentence representation, integrate hierarchical features into its original semantics, and build a representation space with hierarchical semantics of sentences. We design a novel representation of the hierarchical semantics of sentences. Inspired by translation models such as TransE to embed entities and relationships of multi-relational data in low-dimensional vector spaces~\cite{NIPS2013_1cecc7a7}, we obtain more efficient entailment steps based on integrating hierarchical semantics of sentences. 

In this work, we propose an architecture of integrating the Hierarchical Semantics of sentences under the framework of Controller-Generator called \textbf{HiSCG}. We decompose it into three parts: \textbf{hierarchical semantic encoder}, \textbf{selection controller}, and \textbf{intermediate generation}. Firstly, as our main innovative module, the hierarchical semantic encoder adjusts the embedding space to establish hierarchical connections between facts. In an entailment tree, hierarchical connections can be propagated as levels increase. Therefore, all relevant facts form a cluster with the hypothesis as the core, which establishes a hierarchical mapping and increases the retrieval performance of relevant facts accompanied by the gain on step selection. Secondly, the selection controller module selects two facts from the fact set with hierarchical representation for compositions. We retain the previous training methods and add a loss function that can better utilize hierarchical representation. Thirdly, the intermediate generation module based on sequence-to-sequence language models generates conclusions from the two facts (premises), which will be added to the fact set for subsequent selection and generation. In summary, we have three contributions described below. 
\begin{itemize}
	\item[$\bullet$] To the best of our knowledge, we are the first to integrate hierarchical features into the original semantics of sentences, and build a hierarchical representation space to construct entailment trees.
	\item[$\bullet$] Experiments on three settings on the EntailmentBank benchmark  demonstrate that our method achieves comparable results compared to other existing baselines.
	\item[$\bullet$] Further experiments on two cross datasets indicate that our model has better generalization abilities.
\end{itemize}

\section{Related Works}
\textbf{Interpretability of QA Systems} Understanding the chain of reasoning from facts to hypotheses can help to establish an explainable QA system. The approaches based on rationale emphasize paragraph retrieval, including sparse retrieval represented by BM25~\cite{INR-019} and dense retrieval represented by DPR~\cite{karpukhin-etal-2020-dense}, which are mostly under the retriever-reader framework~\cite{izacard2021leveraging, yu-etal-2022-kg}. Instead of simple textual explanations, there have been other attempts to produce a sequence of reasoning steps by iterative retrieval, which can be divided into two subcategories. \textit{Use entity links/hyperlinks} to extract and match keywords, adding hyperlinks and paragraphs with hyperlinks to previous paragraphs~\cite{shao2021memory}. \textit{Reformulate the question} by adding to, modifying, or re-weighting the question texts \cite{10.1145/3404835.3462853, yadav-etal-2021-want, yadav-etal-2020-unsupervised}. 

With the introduction of CoT, the \textit{self-explanation} of language models become a appealing way to gain more insight into predictions. Wei et al.~\cite{NEURIPS2022_9d560961}, Kojima et al.~\cite{NEURIPS2022_8bb0d291} and Lampinen et al.~\cite{lampinen-etal-2022-language} leverage a few training examples as prompts without updating any parameters of the LLMs to generate explanations. However, ChatGPT~\cite{openai_2022}, the comprehensive model with the best performance among them, can produce the hallucination of human-like fluency and confidence without faithfulness to the truth \cite{doi:10.1148/radiol.230163, doi:10.1126/science.adg7879}. In addition, the applied methodologies are still poorly informed by theories, making them still lack credibility and scalability for real-world applications~\cite{guo2023close, davis2023mathematics}. This enables the retrieval-based reasoning to still have priority.

\textbf{Entailment Tree Generation} ~\cite{dalvi-etal-2021-explaining} proposes the EntailmentBank dataset, where the trees are composed of multi-premise textual entailment steps. ~\cite{neves-ribeiro-etal-2022-entailment} proposes an iterative retrieval generative reasoner (IRGR) architecture, which enables the model to use intermediate conclusions to generate step-by-step from textual premises to the hypothesis systematically. \cite{hong-etal-2022-metgen} proposes a module-based framework, in which the controller selects the candidate steps and the individual generator supplies conclusions. ~\cite{liu-etal-2022-rlet} introduces reinforcement learning to generate entailment trees with elaborately designed aligned reward function that is consistent with the evaluation. 

\section{Task Definition}
As shown in Fig.~\ref{question_describe}, we display the reasoning lines in the form of entailment trees. Given a hypothesis and a corpus of facts $\mathcal{C}={s_1,s_2,…,s_n}$ (simple textual sentences) containing relevant facts $s^+$ and irrelevant facts $s^-$. The aim is to determine all relevant facts and combine these facts to generate a valid entailment tree. That is, through reliable compositions and reasonable intermediate conclusions generation, a tree with a root node that is semantically close to the hypothesis will be obtained. Each entailment tree represents as $T=(S,I,P,h)$, where $S$ is the set of leaf nodes ($S\in \mathcal{C}$). $I$ is the set of generated intermediate nodes ($I\notin C$). $P$ is all of the reasoning steps (e.g. $s_1 \wedge s_2 \Rightarrow i$,$s_1$,$s_2\in S,i\in I$). $h$ is the root node of the entailment tree.

\section{Architecture}
\begin{figure}[!t]
	\centering
	\includegraphics[width=3.4in,height=3.8in]{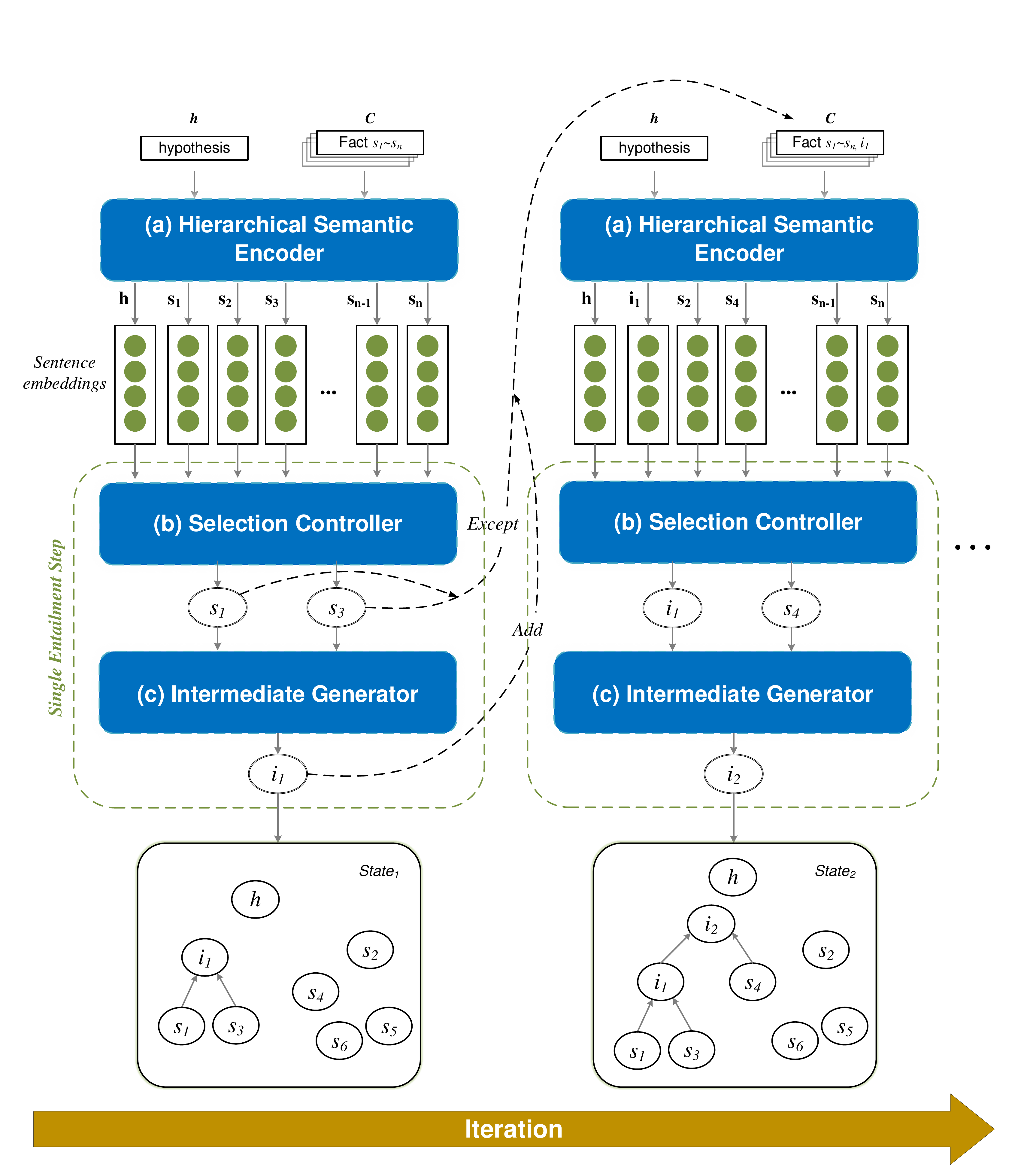}
	\caption {The framework of our proposed model HisCG.} 
	\label{model}
\end{figure}

Following the setting of \cite{bostrom-etal-2021-flexible}, the composition only occurs between two facts. The overall framework of HiSCG contains three modules as shown in Fig.~\ref{model}. We use the hierarchical semantic encoder to obtain hierarchical semantic embeddings of sentences. In each entailment step, the selection controller selects appropriate steps in the fact set. The intermediate generator generates intermediate conclusions for selected steps. We introduce the three modules in detail in Sections \ref{sec4_1}, \ref{sec4_2}, and \ref{sec4_3}, respectively. Section \ref{sec4_4} describes the reasoning strategy of our framework.


\subsection{Hierarchical Semantic Encoder}
\label{sec4_1}
Existing methods directly obtain the embeddings of all facts in $\mathcal{C}$. In this way, when the entailment tree is deep, the relevant fact $s^+$ and irrelevant fact $s^-$ have insignificant differences since the facts have low syntactic and semantic similarity with hypothesis $h$. This will not be conducive to improving the generalization performance of subsequent combination operations. We intuitively think that the problem needs to be solved in the hierarchical semantic space. We propose a novel method to augment the sentence representation, establishing the relationship between the conclusion $i$ and the premises ${s_b,s_e }$. To this end, large quantities of single steps are required. Following \cite{hong-etal-2022-metgen}, we use synthetic data extracted from Wikipedia based on rules as simple samples. 

\begin{figure}[!bht]
	\centering
	\includegraphics[width=3.0in]{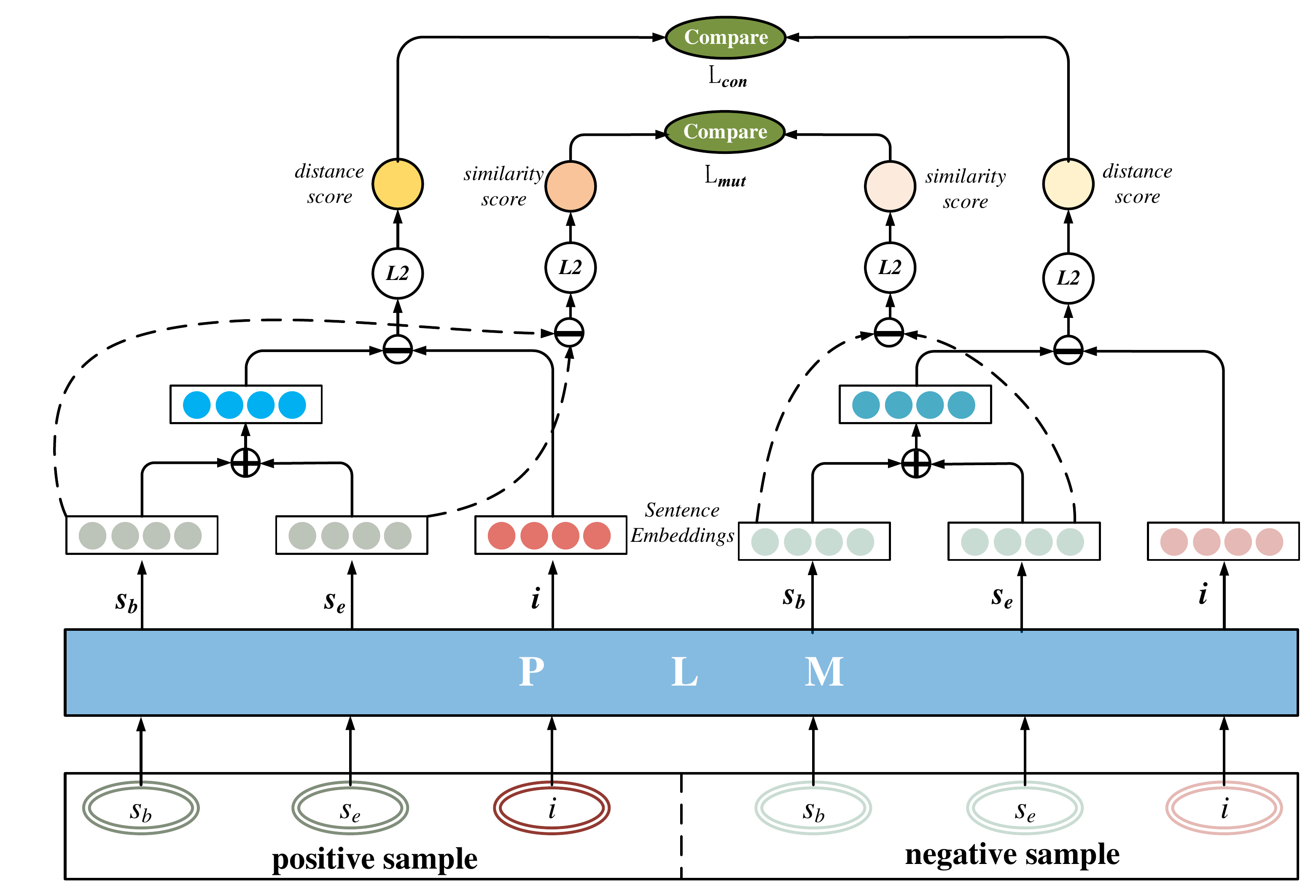}
	\caption {The training details of the Hierarchical Semantic Encoder.} 
	\label{model_encoder}
\end{figure}

Inspired by TransE of embedding entities and relations, we replace those entity and relation nodes with each fact in $\mathcal{C}$. We establish connections between premises and conclusions from the perspective of augmenting hierarchical semantics. For the premises ${s_b,s_e }$ and the conclusion $i$, the embedding of $i$ should be close to the embedding of $s_b$ plus the embedding of $s_e$. As shown in Fig.~\ref{model_encoder}, the premises ${s_b,s_e }$ and the conclusion $i$ will be input into the Pre-trained Language Model (PLM) to compute their sentence embedding. The training objectives are making the golden conclusion and the adduct of $s_e$ and $s_b$ as close as possible, and connecting relative premises simultaneously. 

In terms of implementation, we hope that $s_b$ and $s_e$ are getting closer, and the embedding of $i$ should be close to the embedding of $s_b$ plus the embedding of $s_e$ in the hierarchical representation space $\varepsilon$. According to the vector addition rule, it ultimately displays that $s_b$ and $s_e$ are constantly moving closer to $i$, and $i$ has a geometric diagonal property between them. 

This phenomenon is also reflected in the transmission of correlation with the increase of depths in the entailment tree. Specifically, in a single step $s_{(m,b)}\wedge s_{(m,e)}\Rightarrow i_m$ at \textit{m}-level in the tree, $i_m$ as an intermediate node will become the premise at \textit{m-1}-level (assuming $i_m=s_{(m-1,b)}$). Due to the establishment of the geometric relationship (i.e.,$s_{(m-1,b)}\wedge s_{(m-1,e)}\Rightarrow i_{(m-1)}$), $s_{(m,b)}$ and  $s_{(m,e)}$ are correlated with $i_{(m-1)}$, and the correlations are continuously transmitted up to the root node. The loss function for this stage is shown in Equation~\ref{equ3}.
\begin{equation}
	\label{equ1}
	\begin{split}
		\mathcal{L}_{con} = \sum_{(s_b,s_e,i)\in S}\sum_{S^\prime}[d(s_b+s_e, i)-d(s_b^\prime+s_e^\prime, i)+\gamma_1]_+
	\end{split}
\end{equation}
\begin{equation}
	\label{equ2}
	\begin{aligned}
		\mathcal{L}_{mut}=\sum_{(s_b,s_e,i)\in S}[d(s_b,s_e)-d(s_b,s_e^\prime)+\gamma_2]_+
	\end{aligned}
\end{equation}
\begin{equation}
	\label{equ3}
	\mathcal{L}_{cor}=\mathcal{L}_{con}+\mathcal{L}_{mut}
\end{equation}
Where $[x]_+$ denotes the positive part of $x$, $\gamma_i>0$ is a margin hyperparameter.  $d(s_b+s_e, i)$ is for measuring distance, which we take to be the L2-norm, and $S^\prime=\{(s_b^\prime,s_e,i)\cup(s_b,s_e^\prime,i)\}$, $s^\prime$ is the negative sample of $s$. We select other samples trained in the same batch as negative samples, namely $D_{batch}=\{\left(s_b^1,s_e^1\right),\left(s_b^2,s_e^2\right),$ $\ldots,\left(s_b^M,s_e^M\right)\}$ ($M$ donates the batch size), $s_b^j(j\neq i)$ is the negative sample of $s_b^i$, and $s_e^j(j\neq i)$ is the negative sample of $s_e^i$.

\subsection{Selection Controller}
\label{sec4_2}
There are $C_n^2$ step selections in the given $\mathcal{C}$. We use the selection controller to obtain reasonable steps. A selection controller should serve two purposes. On the one hand, it identifies relevant facts $s^+$ due to distractors contained in $\mathcal{C}$. On the other hand, it determines single entailment steps required at a certain state. We implement both deductive and abductive modules (e.g., $i-s_b \Rightarrow s_e$) to perform forward and backward reasoning respectively~\cite{hong-etal-2022-metgen}. We concatenate the hypothesis and all facts of the tree as :
\begin{equation}
	\begin{split}
		T = {Enc}_{cor}([CLS]\ h \ \{[SEP] \ s_l\}_{n} [SEP],\ l\in[1,n])
	\end{split}
\end{equation}
${Enc}_{cor}$ is the hierarchical semantic encoder. After encoding in hierarchical representation space, the hypothesis $h$ denotes as $\mathbf{h}\in\mathbb{R}^d$, and the factual text $s_l$ denotes as $\mathbf{s}_{\mathbf{l}}\in\mathbb{R}^d$, where $d$ is the vector dimension. The ranking loss is utilized to get all relevant facts. We assume relevant facts closer to the root have higher priority. The binary cross-entropy loss function is to further identify relevant facts from distractors. The loss for scoring facts is:
\begin{equation}
	\label{equ4}
	\begin{split}
		\mathcal{L}_{fact}=&\frac{1}{N_1}\sum_{\mathbf{s}_\mathbf{t}^+,\mathbf{s}_\mathbf{d}^+}{[G_{fact}(\mathbf{s}_\mathbf{t}^+)-G_{fact}(\mathbf{s}_\mathbf{d}^+)+\gamma_3]_+} \\
		&-\lambda\cdot\frac{1}{N_2}\ \sum_{\mathbf{s}_\mathbf{i}}[y_ilog(G_{fact}(\mathbf{s}_\mathbf{i})) \\ &+(1-y_i)log(1-G_{fact}(\mathbf{s}_\mathbf{i}))]
	\end{split}
\end{equation}
\begin{equation}
	G_{fact}(\mathbf{s}_\mathbf{i})=Sigmoid(\mathbf{h};\mathbf{s}_\mathbf{i})
\end{equation}
Where $\mathbf{s}_\mathbf{t}^+$ and $\mathbf{s}_\mathbf{d}^+$ are the embedding of valid facts, but $\mathbf{s}_\mathbf{t}^+$ is closer to the hypothesis root $\mathbf{h}$ than $\mathbf{s}_\mathbf{d}^+$. $N_i$ is the number of corresponding accumulation elements. $\lambda$ is a hyperparameter. ($\cdot$ ; $\cdot$) means concatenation.

Every state in an entailment tree participates in training. In each state, it contains the formed partial tree and the preparing fact set. Except for the golden step $p^+$, other random matches of two sentences are regarded as negative samples $p^-$. In addition, We believe that the intermediate conclusions formed by golden steps have a higher correlation with the hypothesis due to the established relationship between premises and conclusions. Therefore, we also added the correlation loss, which is different from~\cite{hong-etal-2022-metgen}. The loss for scoring steps is:
\begin{equation}
	\label{equ5}
	\begin{split}
		\mathcal{L}_{step}=&\frac{1}{N_3}\sum_{p^+,p^-}{([G_{step}(p^+)-G_{step}(p^-)+\gamma_4]_+} \\
		&+[d(p^+,\mathbf{h})-d(p^-,\mathbf{h})+\gamma_5]_+) 
	\end{split}
\end{equation}
\begin{equation}
	\begin{split}
		G_{step}(p)=G_{step}(\mathbf{s}_\mathbf{i},\mathbf{s}_\mathbf{j})=Softmax(\mathbf{h};\mathbf{s}_\mathbf{i};\mathbf{s}_\mathbf{j})  
	\end{split}
\end{equation}
The hyperparameters $\alpha$ and $\beta$ incorporate two purposes of the selection controller. The loss for the selection controller is:
\begin{equation}
	\mathcal{L}={\alpha\cdot\mathcal{L}}_{step}+\beta\cdot\mathcal{L}_{fact}
\end{equation}

\subsection{Intermediate Generation}
\label{sec4_3}
The selected steps are utilized as the input of the intermediate generation module, and the conclusions derived are the outputs. This generation method can determine the precision of each entailment step, because the intermediate conclusion is only related to the two inputted premises. In contrast, ensemble generations are subjected to reasoning rules~\cite{neves-ribeiro-etal-2022-entailment, dalvi-etal-2021-explaining}, causing significant hallucinations. The premises ${s_b,s_e}$ are inputted in the sequence-to-sequence model in the format of $[CLS][Suffix]\ s_b$ $[SEP]\ s_e\ [SEP]$ where the $[Suffix]$ is "connection:", and the output is the corresponding conclusion $i$. As for abduction, the input format is $[CLS][Suffix]\ s_b(s_e)$ $[SEP]\ i\ [SEP]$ and the output is the $s_e(s_b)$. The synthetic data mentioned in Section \ref{sec4_1} will also participate in training as silver samples. We use a typical auto-regressive language modeling loss to train the module.

\subsection{Reasoning Strategy}
\label{sec4_4}
In the given fact set $\mathcal{C}$, the selection controller obtains ${score}_{fact}$ of each fact in $\mathcal{C}$. The fact will be discarded if ${score}_{fact}<\delta$, where $\delta$ is a threshold. The sub-set of retained facts is called $\phi$. Then the selection controller obtains ${score}_{step}$ of all potential steps. The step with the highest ${score}_{step}$ will be selected, and the intermediate generation module will generate the intermediate conclusion with the selected step. Note that the intermediate conclusion will add to $\phi$, and the facts implicated in steps will withdraw from $\phi$. This process iterates until the reasoning trace to the hypothesis is found. 

\section{Experimental Settings}
\subsection{Dataset}
We evaluate our method on EntailmentBank~\cite{dalvi-etal-2021-explaining}, a benchmark that employs an entailment tree to represent QA explanation processes in the scientific domain. The detailed composition is shown in Table~\ref{table_split}. Three increasingly complex explanation tasks are included in EntailmentBank. Sufficient valid information without distractors is supplied in Task1. 25 facts with distractors are provided in Task2. We need to distinguish the relevant facts and complete the construction. There is no scope limitation in Task3. We need to build the entailment tree from a corpus.

\begin{table}[!htbp]
	\centering
	\caption{The training/dev/testing split of EntailmentBank}
	\setlength{\tabcolsep}{0.55cm} 
	\begin{tabular}{cccc}
		\toprule
		\textbf{Entailment}	& \textbf{\#Train} & \textbf{\#Dev} & \textbf{\#Test} \\
		\midrule
		Tree &	1131 &	187 &	340 \\
		Step &	3476 &	487 &	902 \\
		Leave & 5764 &	816 &	1518 \\
		\bottomrule
	\end{tabular}
	\label{table_split}
\end{table}

\subsection{Evaluation Metrics}
We follow \cite{dalvi-etal-2021-explaining} to evaluate the prediction tree from four dimensions of leaves, steps, and intermediates.

\textbf{Leaves(F1, AllCorrect)}: We compute an \textit{F1} score by comparing predicted leaves and golden leaves. \textit{AllCorrect} =1 only if all predicted leaves conform to golden leaves, and equals 0 otherwise.

\textbf{Steps(F1, AllCorrect)}: To evaluate whether the entailment step is correct. \textit{AllCorrect}=1 if all steps in the prediction tree precisely match the golden tree (i.e., \textit{F1}=1).

\textbf{Intermediates(F1, AllCorrect)}: We use \textit{BLEURT-Large-512}~\cite{sellam-etal-2020-bleurt}
to measure the synthesized intermediate nodes. The predicted conclusion is correct when the bleurt score of the aligned pair exceeds the threshold $\tau$=0.28. \textit{AllCorrect}=1 if all predicted conclusions match (i.e., \textit{F1}=1).

\textbf{Overall(AllCorrect)}: The Overall \textit{AllCorrect} is set to 1 only if the \textit{AllCorrect} of leaves, steps, and intermediate all are 1. Overall is the comprehensive evaluation metric. \textit{AllCorrect}=1 means that the predicted and golden tree are exactly matched, which is the most optimal result.

\subsection{Baselines}
\textbf{EntailmentWriter (2021)}~\cite{dalvi-etal-2021-explaining} generates linearized entailment trees via sequence-to-sequence models with T5-11B and T5-Large versions.

\textbf{METGEN (2022)}~\cite{hong-etal-2022-metgen} is a module-based generative framework that divides the reasoning process into premises retrieval and conclusions generation.

\textbf{IRGR (2022)}~\cite{neves-ribeiro-etal-2022-entailment} generates iterative steps, allowing the model to leverage intermediate conclusions to improve retrieval performance.

\textbf{RLET (2022)}~\cite{liu-etal-2022-rlet} is a reinforcement learning framework based on entailment tree generation, which is trained utilizing the cumulative signals across the whole tree.

\subsection{Implementation Details}
We experimented on four A40 GPUs with 48GB of memory. The hierarchical semantic encoder utilizes \textit{albert-xxlarge-v2} \cite{lan2019albert} as the backbone, and is trained for 20 epochs. The selection controller module reuses the hierarchical semantic encoder and is fine-tuned on downstream tasks. We iteratively generate the entailment tree with all provided facts in Task1. As for Task2, the selection controller filters out the distractors, and the output tree is the best match to the hypothesis in all partial trees. In Task3, we leverage IRGR-retriever to select 50 candidate facts for each hypothesis from the corpus, and then use the hierarchical semantic encoder to refine the fact set within 25 facts (similar to Task2). An important property in Task3 is that hypotheses may entail diverse reasoning lines due to massive facts, so the beam search will be used to expand the tree set. The intermediate generation module is implemented with \textit{T5-large}.

\subsection{Hyperparameters}
In Task1, we set the learning rate to 1e-5, batch size to 16, $\lambda$ to 0, and train for 1000 epochs. In Task2, we set the learning rate to 1e-5, $\lambda$ to 1, batch size to 8, and train for 1500 epochs. ${ \gamma}_i$ is 0.1. The weight constants $\alpha$, $\beta$ are both 1.0. The threshold $\delta$ is 0.001 when inferencing on the Task2 test set. In Task3, we reuse the model of Task2 for inferencing. Besides, we employ beam search to increase the probability of correct reasoning in Task3. We use ${top}_p$ and ${top_{abd}}_p$ to select candidate steps for deductive and abductive reasoning, respectively. In the pilot experiment, setting ${top}_p$ to 0.4, ${top_{abd}}_p$ to 0.1, $\delta$ to 0.1 and beam size to 3 can achieve the best results. $\delta$ is the most influential factor among them. 

\section{Result \& Analysis}
\subsection{Main Result}
As shown in Table~\ref{table_main}, HisCG exposes competitive performance on the Overall \textit{AllCorrect} in Task1 and Task2. Specifically in Task1, the leaves \textit{F1/AllCorrect} is 100\% since no distractors, which is not guaranteed by the linearized generation method. The performance of leaves underpins the advancements of other indicators in Task2. The leaves \textit{AllCorrect} of HisCG outperform other non-generative models. Furthermore, our method is better on all \textit{AllCorrect} metrics when compared to EntailmentWriter with \textit{T5-large}. While using \textit{T5-11B} as its backbone, we also achieve noteworthy results on \textit{AllCorrect} with fewer parameters. Regarding the superior performance of EntailmentWriter with \textit{T5-11B}, we guess the reason is the impressive natural-language understanding capabilities brought by the parameter scale, which makes substantial metric improvements after fine-tuning. However, the sharp decrease in Task3 also indicates that this approach heavily relies on shortcut learning, while without adaptability in open-setting reasoning.

We also exceed all baselines on the strictest Overall \textit{AllCorrect} in Task3. We also achieved the best results on the leave \textit{F1} and step \textit{AllCorrct} metrics, which demonstrates the effectiveness of the proposed method via hierarchical semantic augmented. In the hierarchical embedding space, more relevant facts are retrieved, and the performance of step selection is improved as a by-product due to the established relationship between premises and conclusions. The intermediate \textit{F1/AllCorrect} is greater than that of the baselines, which we believe benefited from the step \textit{AllCorrect} and the data augmentation.
\begin{table*}[!thbp]
	\centering
	\caption{Experiment results on Task1 \& Task2 \& Task3 test set. All baseline results come from published papers.}
	\setlength{\tabcolsep}{0.38cm} 
	\begin{tabular}{c|cccccccc}
		\toprule
		\multirow{2}*{\textbf{Task}} & \multirow{2}*{\textbf{Method}} & \multicolumn{2}{c}{\textbf{Leaves}} & \multicolumn{2}{c}{\textbf{Steps}} & \multicolumn{2}{c}{\textbf{Intermediates}} & \textbf{Overall} \\
		\cmidrule(lr){3-4}\cmidrule(lr){5-6}\cmidrule(lr){7-8}\cmidrule(lr){9-9}
		& & F1 &AllCor. &F1 &AllCor. &F1 &AllCor. &AllCor. \\
		\midrule
		\multirow{6}*{\textbf{Task1}}&EntailmentWriter (T5-11B)&99&89.4&51.5&38.2&71.2&38.5&35.3 \\
		& EntailmentWriter (T5-large)&98.7&86.2&50.5&37.7&67.6&36.2&33.5 \\
		& IRGR (T5-large)&97.6&89.4&50.2&36.8&62.1&31.8&32.4 \\
		& METGEN &100.0 &100.0 &57.9 &42.1 &71.3 &39.2 &37.0 \\
		& RLET&100.0&100.0&54.6&40.7&66.9&36.3&34.8 \\
		&HisCG (ours)&\textbf{100.0}&\textbf{100.0}&\textbf{59.0}&\textbf{44.7}&\textbf{72.2}&\textbf{41.3}&\textbf{38.2} \\
		\midrule
		\multirow{6}*{\textbf{Task2}}&EntailmentWriter(T5-11B)&89.1&48.8&41.4&27.7&\textbf{66.2}&31.5&25.6 \\
		& EntailmentWriter (T5-large)&84.3&35.6&35.5&22.9&61.8&28.5&20.9 \\
		& IRGR (T5-large)&69.9&23.8&30.5&22.4&47.7&26.5&21.8 \\
		& METGEN &83.7 &48.6 &41.7 &30.4 &62.7 &32.7 &28.0 \\
		& RLET&81.0&39&38.5&28.4&56.3&28.6&25.7 \\
		& HisCG (ours)&\textbf{89.9}&\textbf{49.8}&\textbf{43.6}&\textbf{32.2}&64.6&\textbf{33.3}&\textbf{28.8} \\	
		\midrule	
		\multirow{6}*{\textbf{Task3}}&EntailmentWriter(T5-11B)&39.9&3.8&7.4&2.9&35.9&7.1&2.9 \\
		& EntailmentWriter (T5-large)&35.7&2.9&6.1&2.4&33.4&7.7&2.4 \\
		& IRGR (T5-large)&46.6&10.0&11.3&8.2&38.7&20.9&8.2 \\
		& METGEN&34.8&8.7&9.8&8.6&36.6&20.4&8.6 \\
		& RLET&46.2&\textbf{11.4}&\textbf{15.2}&9.6&41.4&17.6&9.4 \\
		&HisCG (ours)&\textbf{47.0}&10.0&14.5&\textbf{9.9}&\textbf{45.0}&\textbf{22.9}&\textbf{9.7} \\	
		\bottomrule
	\end{tabular}
	\label{table_main}
\end{table*}
\begin{table*}[!htbp]
	\centering
	\caption{Ablation results of Task 1 \& Task 2 \& Task 3 on EntailmentBank test set. hse. represents the hierarchical semantic encoder and the correlation loss in equation~\ref{equ5}.}
	\setlength{\tabcolsep}{0.48cm} 
	\begin{tabular}{ccccccccc}
		\toprule
		\multirow{2}*{\textbf{Task}} & \multirow{2}*{\textbf{Method}} & \multicolumn{2}{c}{\textbf{Leaves}} & \multicolumn{2}{c}{\textbf{Steps}} & \multicolumn{2}{c}{\textbf{Intermediates}} & \textbf{Overall} \\
		\cmidrule(lr){3-4}\cmidrule(lr){5-6}\cmidrule(lr){7-8}\cmidrule(lr){9-9}
		& & F1 &AllCor. &F1 &AllCor. &F1 &AllCor. &AllCor. \\
		\midrule
		\multirow{2}*{\textbf{Task1}}&HisCG&\textbf{100.0}&\textbf{100.0}&\textbf{59.0}&\textbf{44.7}&\textbf{72.2}&\textbf{41.3}&\textbf{38.2} \\
		& HisCG w/o hse.&100,0&100.0&57.6&42.1&71.8&39.8&36.8 \\
		\midrule
		\multirow{2}*{\textbf{Task2}}& HisCG&\textbf{89.9}&\textbf{49.8}&\textbf{43.6}&\textbf{32.2}&64.6&\textbf{33.3}&\textbf{28.8} \\
		& HisCG w/o hse.&82.5&38.5&34.0&22.6&56.5&28.5&19.4 \\
		\midrule
		\multirow{2}*{\textbf{Task3}}&HisCG&\textbf{47.0}&\textbf{10.0}&\textbf{14.5}&\textbf{9.9}&\textbf{45.0}&\textbf{22.9}&\textbf{9.7} \\
		& HisCG w/o hse.&42.8&7.1&9.2&7.1&43.6&21.6&6.5 \\
		\bottomrule
	\end{tabular}
	\label{abation_cor}
\end{table*}
\begin{table*}[!htbp]
	\centering
	\caption{Ablation results on task1 \& task2 \& task3 in the intermediate generation module under different settings. w/rule means rule-based method, and w/o syn. means without using synthetic data during training.}
	\setlength{\tabcolsep}{0.48cm} 
	\begin{tabular}{ccccccccc}
		\toprule
		\multirow{2}*{\textbf{Task}} & \multirow{2}*{\textbf{Method}} & \multicolumn{2}{c}{\textbf{Leaves}} & \multicolumn{2}{c}{\textbf{Steps}} & \multicolumn{2}{c}{\textbf{Intermediates}} & \textbf{Overall} \\
		\cmidrule(lr){3-4}\cmidrule(lr){5-6}\cmidrule(lr){7-8}\cmidrule(lr){9-9}
		& & F1 &AllCor. &F1 &AllCor. &F1 &AllCor. &AllCor. \\
		\midrule
		\multirow{2}*{\textbf{Task1}}&HisCG&\textbf{100.0}&\textbf{100.0}&\textbf{59.0}&\textbf{44.7}&\textbf{72.2}&\textbf{41.3}&\textbf{38.2} \\
		& HisCG w/o syn.&100.0&100.0&58.7&44.6&71.4&41.2&37.9 \\
		\midrule
		\multirow{2}*{\textbf{Task2}}&HisCG&\textbf{89.9}&\textbf{49.8}&\textbf{43.6}&\textbf{32.2}&\textbf{64.6}&\textbf{33.3}&\textbf{28.8} \\
		& HisCG w/o syn.&87.2&48.7&38.7&29.9&62.4&32.7&28.3 \\
		\midrule
		\multirow{2}*{\textbf{Task3}}&HisCG&\textbf{47.0}&\textbf{10.0}&\textbf{14.5}&\textbf{9.9}&\textbf{45.0}&\textbf{22.9}&\textbf{9.7} \\
		& HisCG w/o syn.&46.2&10.0&14.2&9.9&42.7&20.9&9.7 \\
		\bottomrule
	\end{tabular}
	\label{abation_extra}
\end{table*}

\subsection{Ablation Result}
\begin{figure}[!t]
	\centering
	\includegraphics[width=3.4in]{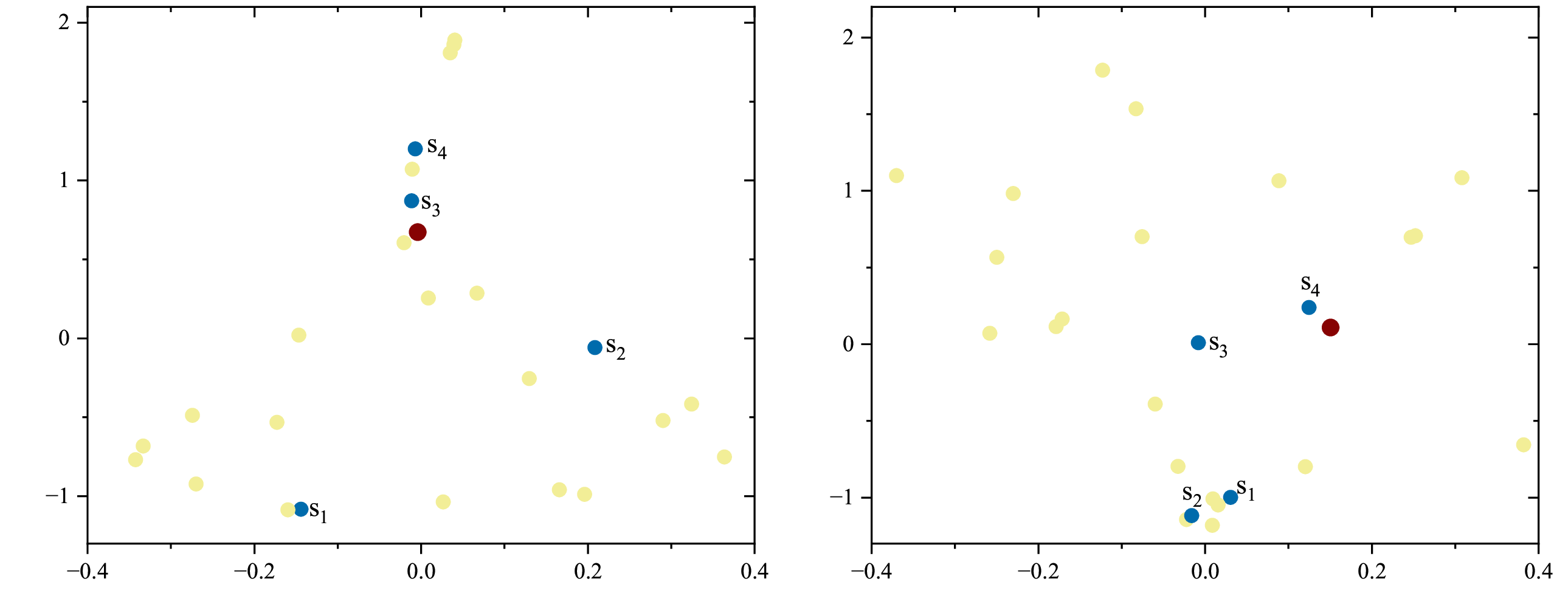}
	\caption {Use t-sne to visualize the example in Fig \ref{question_describe}. Red nodes represent hypotheses, blue nodes are relevant facts, and yellow nodes are distractors.The representation distributions are from  w/o hierarchical semantics encoder (left) and w hierarchical semantics encoder (right).} 
	\label{fig_distribution}
\end{figure}

\textbf{Effectiveness of hierarchical semantic encoder} With the desire to verify whether the hierarchical semantic encoder works, we performed ablation studies on three tasks. As shown in Table~\ref{abation_cor}, without the hierarchical semantic encoder in Task2, the decline of leaves \textit{F1/AllCorrect} affects the sharp decrease of other metrics, which reveals our effective retrieval strategy. The leaves/steps/intermediates/overall \textit{AllCorrect} also relatively fall by 29\%/28\%/6\%/33\% in Task3. This demonstrates that our strategy has established profound connections between sentences, enabling step selection more robust. Additionally, we use t-SNE to visualize the representation of facts and hypotheses in Fig. \ref{fig_distribution}. As the picture shows, the hierarchical semantic encoder can gather relevant facts near the hypothesis and distance irrelevant facts. And overall, it brings sentences with combinatorial relationships closer together. This demonstrates that HisCG can capture the hierarchical semantics required for sentences in the entailment tree construction, which provides good prerequisites for the controller to improve model performance.

\textbf{Efficiency of Intermediate Generation} We further investigate whether redundant training data has a prominent influence on the intermediate generation module. We conduct ablation studies on Task1\&Task2\&Task3. The results in Table~\ref{abation_extra} show that extra training data has brought certain advancements, but it is not the primary driver of Overall \textit{AllCorrect}.

\subsection{Results Breakdown}
\begin{figure}[!t]
	\centering
	\includegraphics[width=2.8in,height=2.0in]{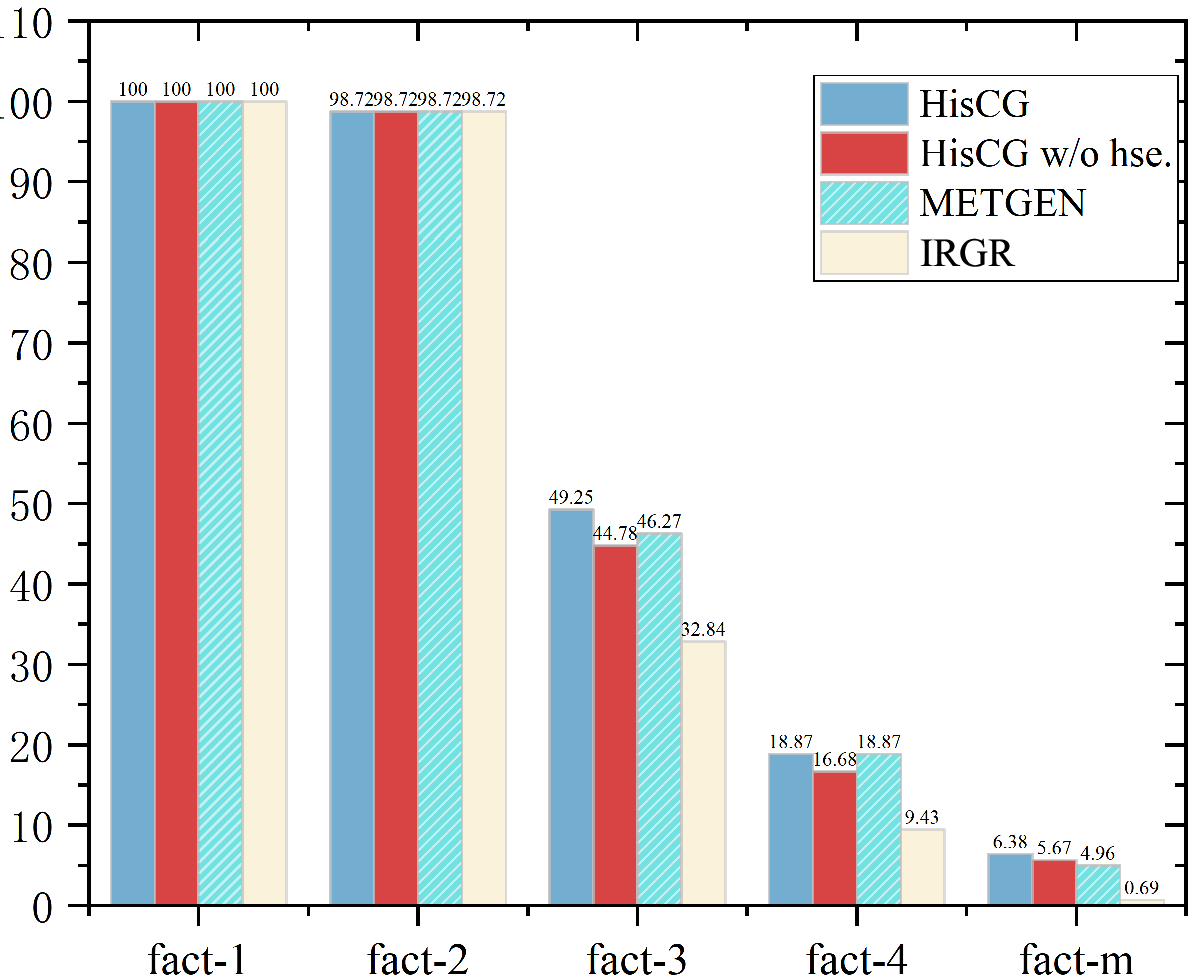}
	\caption {The result of Task1 test set in terms of the number of leaves. The value on bars indicates the proportion of entailment trees containing a certain number of leaves that are correctly predicted by HisCG.} 
	\label{fig_analyze}
\end{figure}

We analyze Task1 in terms of the number of leaves to better assess the contribution of the hierarchical semantic module. As shown in Fig.~\ref{fig_analyze}, HisCG can predict all correct steps due to the uniqueness (accounting for 22.9\%) in the setting of fact-1/fact-2. For the case of fact-3 (accounting for 19.7\%), HisCG achieves significant improvements. When the number of leaves is greater than 4, HisCG comes with relatively poor performance. On the one hand, the increase of relevant facts introduces multiplied exponentially difficulty. On the other hand, some reasoning steps are imprecise in the case of increasing relevant facts, which brings noisy interference to the model.

\subsection{Entailment Step Error Analysis}
\begin{figure}[!t]
	\centering
	\includegraphics[width=2.5in,height=1.5in]{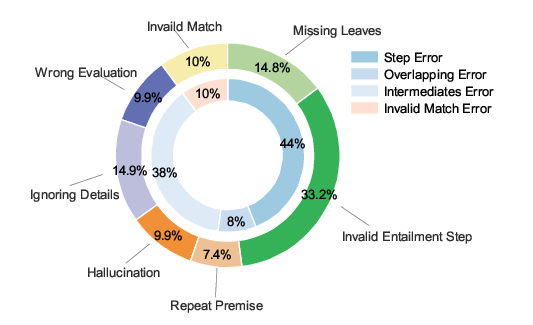}
	\caption {Entailment tree error analysis. The inner ring describes the proportion of major errors. Overlapping Error means that there are both Step Error and Intermediates Error. The outer ring depicts the subdivision subcategories corresponding to the inner ring.} 
	\label{fig_error}
\end{figure}

We analyze the errors manually to further understand the strengths and weaknesses of HisCG in the test set of Task2. Fig.\ref{fig_error} illustrates the error categories comprehensively.

\begin{figure}[bhtp]
	\centering
	\includegraphics[width=3.3in]{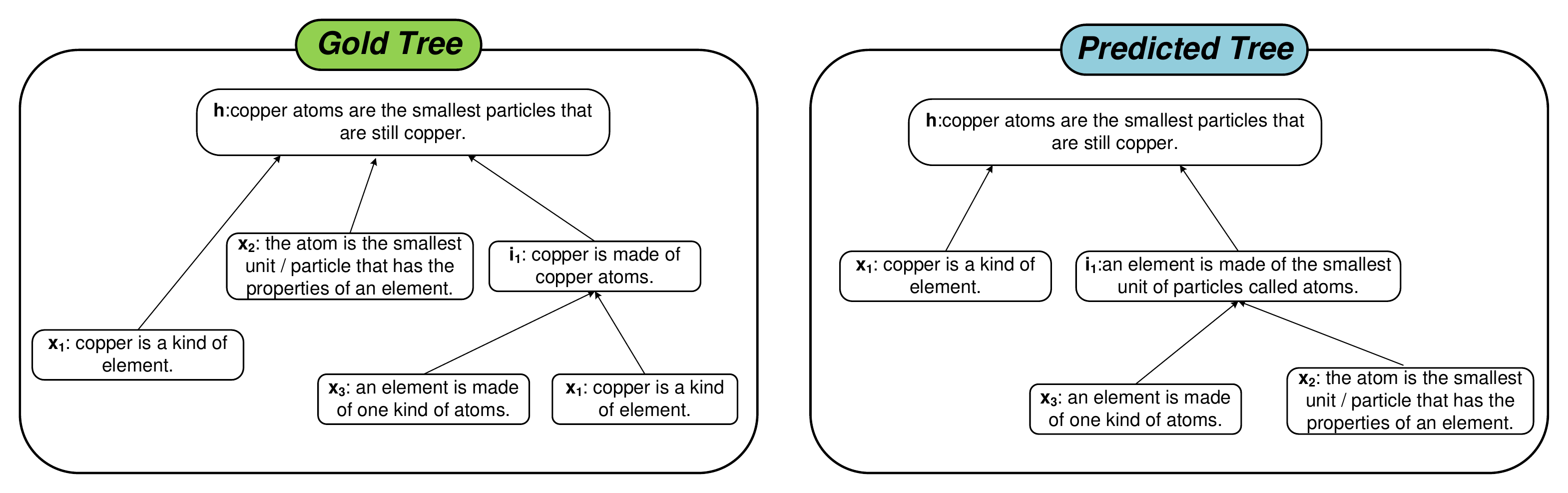}
	\caption {An example of wrong evaluation.} 
	\label{example_evaluation}
\end{figure}
There are three major categories of errors. The \textbf{step errors} account for 52\%, and the \textbf{intermediate errors} account for 46\% (existing cross-cases). Moreover, there are cases of underestimation (10\%) in our manual evaluation (\textbf{match errrors}). One example in the Task2 test set is depicted in Fig.\ref{example_evaluation}.

Steps error includes \textbf{missing necessary leaves} and \textbf{selecting invalid steps} under complete necessary premises. Intermediates error includes \textbf{repeated premises} (outputs consistent with the inputs), ignored important details (i.e., critical connection details) and \textbf{generated hallucinatory conclusions} caused by the probability distribution of words.

\subsection{Cross-dataset Setting}
Annotated reasoning traces are difficult to expand, making generalization ability even more imperative. We follow \cite{hong-etal-2022-metgen} to conduct generalization experiments on the eOBQA dataset and the eQASC dataset respectively, which can be regarded as a single-step entailment tree, that is, $s_1$ and $s_2$ from the candidate sentences can form as $s_1 \wedge s_2 \Rightarrow h$. We select applicable cases and apply them to the Task2 model. P@1 and NDCG will be exerted to evaluate the model \cite{jhamtani-clark-2020-learning}. The results in Table~\ref{table_cross} show nearly 6\% and 5\% improvements on the P@1, which demonstrates that our proposed model has better generalization ability in the single-step entailments.
\begin{table}[!htbp]
	\centering
	\caption{Cross-dataset results on the eQASC and eOBQA test split.}
	\setlength{\tabcolsep}{0.25cm} 
	\small
	\begin{tabular}{ccccc}
		\toprule
		\multirow{2}*{\textbf{Method}} & \multicolumn{2}{c}{eQASC} & \multicolumn{2}{c}{eOBQA} \\
		\cmidrule(lr){2-3}\cmidrule(lr){4-5}
		& P@1&NDCG&P@1&NDCG \\
		\midrule
		\small{\makecell[c]{EntailmentWriter }}&52.48&73.14&69.07&89.05 \\
		\small{\makecell[c]{EntailmentWriter-Iter}}&52.56&73.28&72.15&90.19 \\
		METGEN&55.81&74.19&74.89&90.50 \\
		HisCG&\textbf{62.04}&\textbf{78.14}&\textbf{77.22}&\textbf{91.59} \\
		\bottomrule
	\end{tabular}
	\label{table_cross}
\end{table}

\section{Conclusion}
We propose a method for integrating the hierarchical semantics of sentences to generate explanatory entailment trees iteratively, which consists of the hierarchical semantic encoder, selection controller, and intermediate generator. In light of the representation of entities and relation nodes in translate models, we innovatively designed the hierarchical semantic encoder with the connection between the premise and the conclusion as the intention. Experimental results compared with other baselines demonstrate that our architecture has comparable advantages. From the above phenomena, it can be seen that retrieval and reasoning performance is still an essential bottleneck. We will investigate efficient paradigms based on semantic features more thoroughly in the future.

\bibliographystyle{named}
\bibliography{ijcai24}

\end{document}